\documentclass[a4paper, 10pt, conference]{ieeeconf}      
\usepackage{FG2025}

\FGfinalcopy 
\usepackage{svg}
\usepackage{amsmath,amssymb}
\usepackage{url}
\usepackage{listings}
\usepackage{multirow}

\IEEEoverridecommandlockouts                              
\def\etal{\emph{et al}. }

\title{\LARGE \bf
Robust Deepfake Detection for Electronic Know Your Customer Systems Using Registered Images
}

\author{\parbox{16cm}{\centering
    {\large Takuma Amada$^1$, Kazuya Kakizaki$^1$, Taiki Miyagawa$^1$, Akinori F. Ebihara$^1$,\\Kaede Shiohara$^2$, and Toshihiko Yamasaki$^2$}\\
    {\normalsize
    $^1$ NEC Corporation\\
    $^2$ The University of Tokyo}}
}

\usepackage{fancyhdr}
\begin{document}

\ifFGfinal
\thispagestyle{empty}
\pagestyle{empty}
\else
\author{Anonymous FG2025 submission\\ Paper ID \FGPaperID \\}
\pagestyle{plain}
\fi
\maketitle

\begin{abstract}
In this paper, we present a deepfake detection algorithm specifically designed for electronic Know Your Customer (eKYC) systems.
To ensure the reliability of eKYC systems against deepfake attacks, it is essential to develop a robust deepfake detector capable of identifying both face swapping and face reenactment, while also being robust to image degradation.
We address these challenges through three key contributions:
(1)~Our approach evaluates the video's authenticity by detecting temporal inconsistencies in identity vectors extracted by face recognition models, leading to comprehensive detection of both face swapping and face reenactment. 
(2)~In addition to processing video input, the algorithm utilizes a registered image (assumed to be genuine) to calculate identity discrepancies between the input video and the registered image, significantly improving detection accuracy.
(3)~We find that employing a face feature extractor trained on a larger dataset enhances both detection performance and robustness against image degradation.
Our experimental results show that our proposed method accurately detects both face swapping and face reenactment comprehensively and is robust against various forms of unseen image degradation.
Our source code is publicly available \url{https://github.com/TaikiMiyagawa/DeepfakeDetection4eKYC}.

\end{abstract}

\section{Introduction}

\addtocounter{footnote}{-1} 
\renewcommand{\thefootnote}{}
\footnotetext{\textcopyright 2025 IEEE. Personal use of this material is permitted.  Permission from IEEE must be obtained for all other uses, in any current or future media, including reprinting/republishing this material for advertising or promotional purposes, creating new collective works, for resale or redistribution to servers or lists, or reuse of any copyrighted component of this work in other works.\\  DOI: TBD.}
\renewcommand{\thefootnote}{\arabic{footnote}}

\begin{figure}[t]
\begin{center}
\includegraphics[width=\columnwidth]{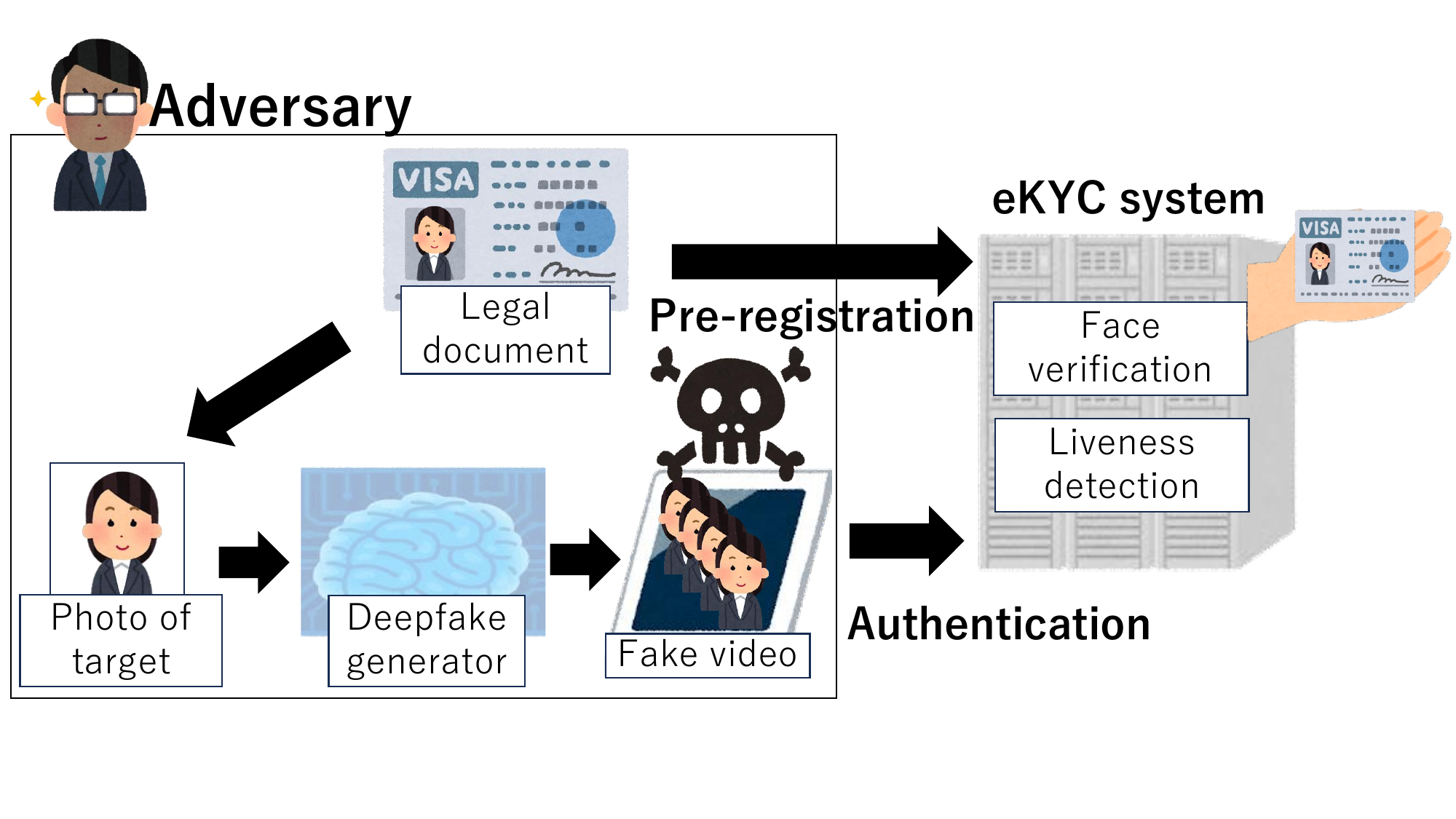}
\end{center}
\caption{Impersonation attack using a deepfake against an eKYC system. The system authenticates identities by comparing facial images from legal documents and videos captured on devices.}
\label{fig:short}
\end{figure}

The advancement of deepfake technology has become a significant threat to electronic Know Your Customer (eKYC) systems~\cite{felouat2024ekyc, do2021potential,do2022potential}, which are foundational to the security protocols in financial institutions (Fig.~\ref{fig:short}).
This threat is exacerbated by increasingly sophisticated deepfake generation algorithms.
Deepfake generation algorithms can be categorized into \textit{face swapping}~\cite{bitouk2008face,korshunova2017fast,li2019faceshifter,DBLP:conf/iccv/NirkinKH19,li20233d,shiohara2023blendface}  and \textit{face reenactment}~\cite{thies2016face2face,kim2018deep,thies2019deferred,zhao2022thin,zhang2023sadtalker}. 
Face swapping involves replacing the face in a source image with one from a target image, while face reenactment alters the facial expressions in the source image to match those of the target image. 

Specialized deepfake detectors for eKYC systems encounter two primary challenges: comprehensive detection of various generation methods of deepfakes and maintaining robustness against image degradation.
These two challenges are attributed to adversaries' ability to exploit their own devices to deceive eKYC systems (Fig.~\ref{fig:short}).
Consequently, attackers can use a range of deepfake generation methods, such as face swapping and face reenactment, necessitating deepfake detectors to effectively identify both techniques.
Moreover, the susceptibility of deepfake detectors to image degradation, such as blurring, additive noise, and compression, makes the issue more complicated. 
This challenge of robustness to image degradation is critical in the context of eKYC systems, where facial images are often captured under diverse conditions and with varying device quality. 
Collectively, these two challenges pose serious security risks to eKYC systems, threatening their reliability and efficacy.

To address these challenges, we propose a deepfake detector consisting of three key components.
(1)~Our model evaluates the video's authenticity by capturing temporal inconsistencies in the identity vectors extracted by the face recognition models, leading to comprehensive detection of both face swapping and face reenactment. 
(2)~It processes both a video and a registered image (assumed to be authentic), calculating identity inconsistencies between the two inputs to enhance detection accuracy.
(3)~We incorporate a high-performance face feature extractor that demonstrates robustness against image degradation, ensuring reliable performance across diverse input conditions.

Our in-dataset evaluation and cross-dataset evaluation show that our approach outperforms a previous baseline~\cite{DBLP:conf/wacv/LiuLDZY23} by 1.5\% and 21.5\% points, respectively. 
Both methods are trained on the Korean DeepFake Detection Dataset (KoDF)~\cite{DBLP:conf/iccv/KwonYNPC21} and evaluated on the Celeb-DF v2 dataset~\cite{DBLP:conf/cvpr/LiYSQL20} in cross-dataset evaluation.
In the robustness evaluation, we compare the decline in AUCs when Gaussian blur is applied to input images. 
Our approach mitigates the decline in AUCs, achieving improvements over the video-level baseline~\cite{haliassos2021lips} and the frame-level baseline~\cite{DBLP:conf/cvpr/ShioharaY22} by 29.7\% and 24.9\%, respectively.
Additionally, we experimentally show that our method is more robust to image degradation than other baselines.
Furthermore, we demonstrate that a higher-performance, pre-trained face recognition model enhances both detection performance and robustness against image degradation. 
Therefore, the design principle of leveraging identity vector dynamics extracted by a large-scale pre-trained model is likely to benefit the deepfake detection community.

\section{Related Work}
\label{sec:related_work}
\subsection{Deepfake Generation}
Two primary methods of creating deepfakes are: \textit{face swapping}~\cite{bitouk2008face,korshunova2017fast,li2019faceshifter,DBLP:conf/iccv/NirkinKH19,li20233d,shiohara2023blendface}  and \textit{face reenactment}~\cite{thies2016face2face,kim2018deep,thies2019deferred,zhao2022thin,zhang2023sadtalker}. 
Face swapping aims to replace the source person's face with the target person's.
Since face swapping usually renders only the region of the source person's face in the target person's image, blending boundaries are left as visual artifacts due to inconsistencies in the colors and textures between the source and the target face.
Face reenactment aims to transform the facial expressions and actions of the source person into those of the target person. 
It manipulates only the facial expressions and actions of the source person. Therefore, the identity is preserved before and after the manipulation.

\subsection{Deepfake Detection}
Current deepfake detection models employ neural networks such as shallow networks~\cite{afchar2018mesonet}, capsule networks~\cite{nguyen2019use}, and recurrent convolutional networks~\cite{guera2018deepfake, sabir2019recurrent}, which discern spatial artifacts and temporal inconsistencies within fake images. 
One common technique for detecting face swapping is identifying inconsistencies in the blending of facial features with the background, which frequently arise during image manipulation~\cite{afchar2018mesonet,nguyen2019use,guera2018deepfake, sabir2019recurrent,li2020face}. 
Face X-ray~\cite{li2020face} represents the boundary between the face and background images. 
Pair-wise self-consistency learning~\cite{zhao2021learning} detects inconsistencies between face and background by dividing a single image into patches and measuring patch-wise similarity. 
Detectors trained on self-blended images (SBIs), artificial composites of source and target images created from a single image, have demonstrated superior detection performance for unseen manipulations~\cite{DBLP:conf/cvpr/ShioharaY22}.
Yan \etal introduce latent space data augmentation (LSDA), which enables the model to detect a wider range of forgeries without overfitting to specific artifacts and to enhance generalization by augmenting the forgery feature in latent space~\cite{yan2024transcending}.
For face reenactment detection, leveraging temporal patterns across video frames has proven effective~\cite{amerini2019deepfake, cozzolino2021id}.
LipForensics~\cite{haliassos2021lips} trains a time-series model based on the lip reading model to assess the authenticity of videos by focusing on mouth movements. 
Amerini \etal show that inter-frame dissimilarity using optical flow is a clue to distinguish face reenactment from the pristine videos~\cite{amerini2019deepfake}.
Wang \etal propose a training strategy for 3D convolutional networks to capture both spatial and temporal artifacts, improving the generalization of deepfake detectors~\cite{wang2023altfreezing}.

Another successful approach to deepfake detection includes the use of identity vectors extracted by a face recognition model. 
Ramachandran \etal demonstrate how employing face recognition models for deepfake detection fosters generalizability~\cite{ramachandran2021experimental}. 
The approach using identity vectors performs well on face swapping but struggles with face reenactment. 
Cozzolino \etal propose ID-Reveal, which is a mechanism for learning temporal facial features, such as person-specific speech movement patterns, via auxiliary videos of the same ID as the input video~\cite{cozzolino2021id}. 
Dong \etal suggest an identity consistency transformer (ICT) focused on identifying inconsistencies within the face's internal and external regions for face swapping detection, displaying impressive performance even under various image degradation scenarios~\cite{dong2022protecting}. 
Liu \etal propose the Temporal Identity Inconsistency Network (TI$^2$Net), which focuses on the temporal consistency of identity vectors extracted from face recognition models~\cite{DBLP:conf/wacv/LiuLDZY23}.
TI$^2$Net demonstrates strong generalization abilities across datasets with unknown identities and maintains robustness against compression and additive noise.
Huang \etal focus on facial identity in face swapping and detect inconsistency between explicit and implicit identity using a face recognition model~\cite{huang2023implicit}.

\section{Proposed Method}

\begin{figure*}[t]
\begin{center}
\includegraphics[width=1.7\columnwidth]{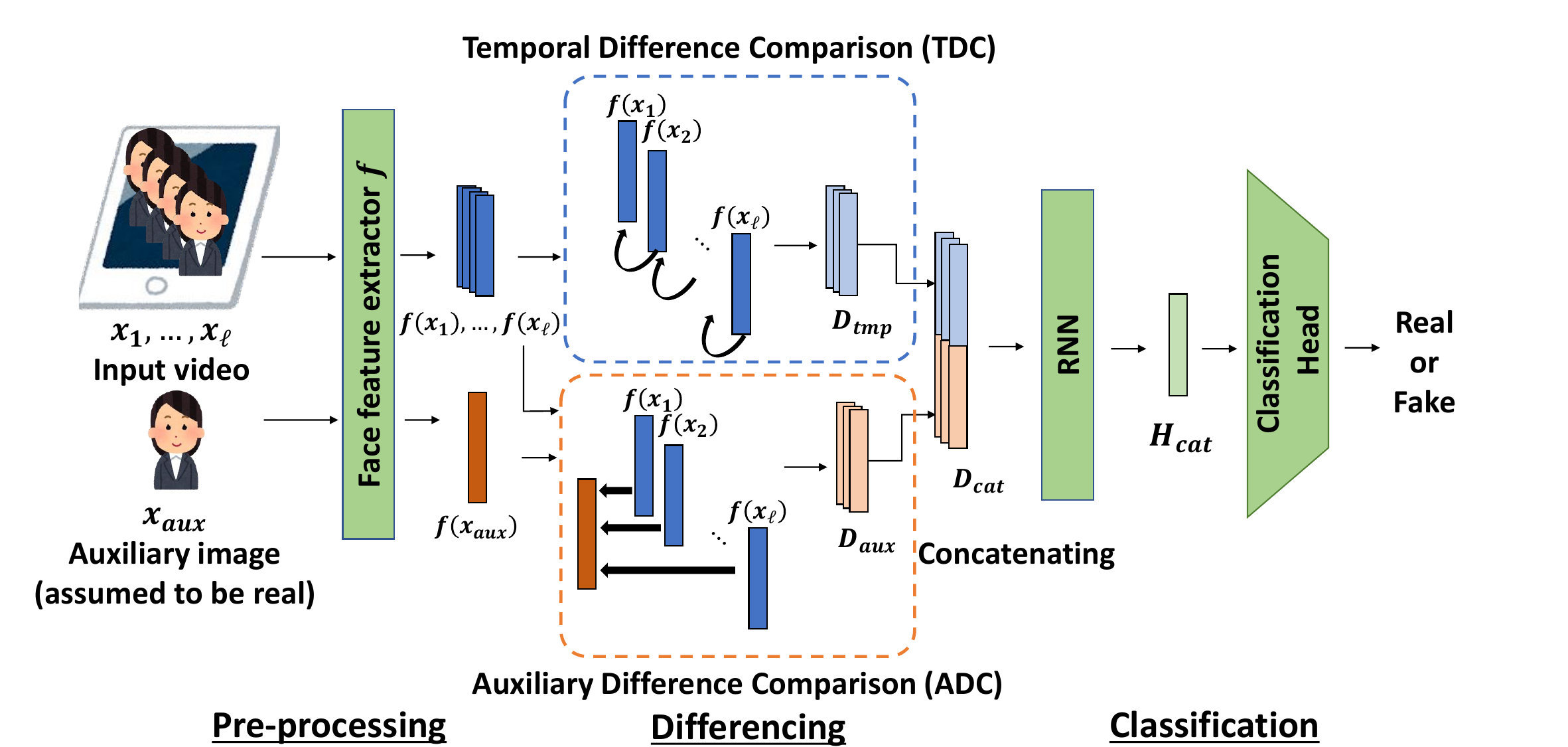}
\end{center}
   \caption{Proposed deepfake detector. It consists of (1) Auxiliary Difference Comparison (ADC), (2) Temporal Difference Comparison (TDC), and (3) a robust feature extractor.}
\label{fig:proposed_method}
\end{figure*}

\subsection{Notation}%
We first provide our notations to describe the eKYC systems and their adversaries.
The eKYC systems internally execute face verification and liveness detection (see Fig.~\ref{fig:short}).
Let $X$ be a set of images with height $H\in \mathbb{N}$, width $W \in \mathbb{N}$, and the number of channels $C \in \mathbb{N}$. Let $V$ be a sequence of images $V:= \{x_{t}\}^{\ell}_{t=1}$, where $x_t \in X$ and $\ell \in \mathbb{N}$ is the length of sequence.
Let $f:\mathbb{R}^{H\times W\times C}\rightarrow \mathbb{R}^{d}$ be a feature extractor, where $d\in\mathbb{N}$ is the feature dimension.
We define face verification as a function \textbf{FV} that maps two images $x_1, x_2 \in X$ to a binary set \{Verified, NotVerified\} as $\textbf{FV}(x_1, x_2)$, and liveness detection \textbf{LD} which maps from a sequence of images $V$ to a binary set \{Detected, NotDetected\} as $\textbf{LD}(V)$.
If \textbf{FV} is Verified and \textbf{LD} is Detected, the eKYC systems ensure that the person providing the documentation is indeed the rightful owner of the identity.
An adversary attempts to input a deepfake video into the eKYC systems by utilizing either face swapping or face reenactment, generating the video in real-time (Fig.~\ref{fig:short}). 
We assume that the adversary, associated with identity $i$, has acquired a document containing the face image $x^j$ of identity $j$, the victim of the impersonation. 
Using a deepfake generator $G$ and $x^j$, the adversary creates a sequence of images $G(x^j)$ that visually resemble the identity $j$ in real-time and submits them to the eKYC systems. 
The eKYC systems verify \textbf{FV} and \textbf{LD} using $x^j$ and $V^j$. 
The result is \textbf{FV} = Verified, as $G(x^j)$ and $x^j$ appear to correspond to the same identity, and \textbf{LD} = Detected, since $G(x^j)$ can replicate any required facial movements in real-time.

\subsection{Proposed Model}
It is essential to accurately detect deepfakes arising from both face swapping and face reenactment, and the detection method must be robust to video quality degradation, as the adversary can use any camera device under various shooting conditions, resulting in a wide range of video qualities.
To achieve (1) improved detection performance, (2) detection of both face swapping and face reenactment, and (3) robustness to video quality degradation, our method consists of (1) Auxiliary Difference Comparison (ADC), (2) Temporal Difference Comparison (TDC), and (3) a robust feature extractor (see Fig.~\ref{fig:proposed_method}).

\subsubsection{Auxiliary difference comparison (ADC)}
ADC enhances deepfake detection by leveraging the registered image, an approach not previously addressed in the literature~\cite{detect_deepfake_on_ekyc,detect_deepfake_on_ekyc2}. 
ADC takes the identity vectors of the registered image and those of video frames, and outputs a sequence of their difference vectors between them.
Specifically, given a registered image $x^{i}_{\rm aux}$ of an identity $i$ and an input video $V^i=\{x_t^i\}^{\ell}_{t=1}$ representing the identity, ADC calculates:
\begin{eqnarray}
\begin{split}
D_{\rm aux} =  \{f(x_1^i)-f(x_{\rm aux}^{i}), f(x_2^i)-f(x_{\rm aux}^i), \\\ldots ,f(x_{\ell-1}^i)-f(x_{\rm aux}^i) \} \, ,
\end{split}
\end{eqnarray}
where $f$ is a face feature extractor.
Note that the registered images are assumed to be real.
This sequence of difference vectors captures identity inconsistency between the registered image and the input video and can be used to detect instantaneous fluctuations of identity introduced by deepfake algorithms.

\subsubsection{Temporal difference comparison (TDC)}
To detect both face swapping and face reenactment, TDC focuses on temporal inconsistency rather than spatial inconsistency because the latter can vary depending on the generation methods of deepfakes.
Following the approach in~\cite{DBLP:conf/wacv/LiuLDZY23}, we compute the following temporal difference vectors:
\begin{eqnarray}
\begin{split}
D_{\rm tmp}  = \{f(x_2^j)-f(x_1^j), f(x_3^j)-f(x_2^j), \\\ldots ,f(x_{\ell}^j)-f(x_{\ell-1}^j) \}.
\end{split}
\end{eqnarray}
Both $D_{\rm aux}$ and $D_{\rm tmp}$ are then fed into a Recurrent Neural Network (RNN), which is subsequently used for classification of the input video (Fig.~\ref{fig:proposed_method}).

\subsubsection{Classification and loss functions}
We further describe the architecture of our model in detail.
To generate input vectors to recurrent neural networks (RNNs), we concatenate features derived from TDC and ADC across the time series as follows:
\begin{eqnarray}
D_{\rm cat} =  D_{\rm tmp} \cup D_{\rm aux}.
\end{eqnarray}
Note that the last element of $D_{\rm cat}$ is truncated to match the number of elements in $D_{\rm tmp}$ and $D_{\rm aux}$. 
The RNN generates $d$-dimension embeddings $H \in \mathbb{R}^{d}$ as temporal features from the last time step. 
These embeddings $H$ are fed into a classification head comprising two fully connected layers. 

We use three standard loss functions: cross-entropy loss for classification, triplet loss to optimize feature embedding, and anchor-positive loss ($L^2$-distance between the anchor and positive) to accelerate training.
The triplet loss comprises anchor, positive, and negative sequences~\cite{DBLP:conf/bmvc/BalntasRPM16}. 
An anchor sequence and a positive sequence are sampled from real videos, and a negative sequence is sampled from fake videos. 
Embeddings $H_{a}, H_{p}, H_{n}$ from these sequences are extracted as described previously. 
We optimize the parameters of the RNNs by minimizing the triplet loss, which pulls the anchor and positive embeddings closer to each other and pushes the anchor and negative embeddings farther apart. The triplet loss is defined as: 
\begin{eqnarray}
L_{\rm tri} = \max(\|H_a-H_p\|_2 - \|H_a-H_n\|_2 + \alpha, 0),
\end{eqnarray}
where $\|\cdot\|$ is the $L^2$-norm and $\alpha$ is the margin, a positive constant. 
We also employ the anchor-positive regularizer to help the convergence of the training loss: 
\begin{eqnarray}
L_{\rm ap} = \|H_a-H_p\|_2 .
\end{eqnarray}
The classification loss is calculated as the cross-entropy loss between the predicted output and the true labels: 
\begin{eqnarray}
L_{\rm cls} = CE(y,p),
\end{eqnarray}
where $CE$ is cross-entropy loss, $y$ is the label of the input, and $p$ is the predicted probability.
Then the overall loss is defined as:
\begin{eqnarray}
L = L_{\rm cls} + \lambda_1 L_{\rm tri} + \lambda_2 L_{\rm ap},
\end{eqnarray}
where $\lambda_1$ and $\lambda_2$ are hyper-parameters that control the weights of loss terms.

\subsubsection{Robust feature extractor}
A notable advantage of our approach is its flexibility to integrate with a variety of feature extractors.
We observed significant improvements in detection accuracy and robustness against image degradation when employing a robust, high-performance face feature extractor, as detailed in Section~\ref{sec:var_feature_extractors}.
In our experiments, we demonstrate that combining this robust face feature extractor with ADC and TDC leads to synergistic improvements in both detection accuracy and robustness against image degradation.

\section{Experiments}
\subsection{Experimental Settings}
\subsubsection{Datasets}
Our method requires an auxiliary image of the same person as the input video, but in eKYC applications, the input video and an auxiliary image often differ visually. 
For instance, the input video is captured in real-time during authentication, whereas the auxiliary image is a photo printed on an identity document.
To meet this requirement, we employ a dataset where auxiliary images are sampled from different videos than the input video, ensuring consistency during training and evaluation.
Despite its prevalence in deepfake-related studies, FaceForensics++ (FF++)~\cite{ff++} cannot be employed for our training and evaluation because it is limited to no more than two videos per identity, making auxiliary image sampling impossible.

Instead, we adopt the Korean DeepFake Detection Dataset (\textbf{KoDF})~\cite{DBLP:conf/iccv/KwonYNPC21} to train models. 
The KoDF dataset contains 62,166 real videos and 175,776 deepfake videos from 403 identities, where each identity includes multiple videos, allowing us to sample an auxiliary image for each video.

The deepfake videos in KoDF consist of five synthesis methods, including FaceSwap (\textbf{FS}) \footnote{GitHub repository of deepfake\_faceswap: \url{https://github.com/deepfakes/faceswap}}, DeepFaceLab (\textbf{DFL})~\cite{DBLP:journals/pr/LiuPGCZZ23}, Face Swapping GAN (\textbf{FSGAN})~\cite{DBLP:conf/iccv/NirkinKH19}, First Order Motion Model (\textbf{FOMM})~\cite{DBLP:conf/nips/SiarohinLT0S19}, and Audio-driven (\textbf{AD})~\cite{DBLP:journals/corr/abs-2002-10137, DBLP:conf/mm/PrajwalMNJ20}). 
\textbf{FS}, \textbf{DFL}, and \textbf{FSGAN} are face swapping manipulations and \textbf{FOMM} and \textbf{AD} are face reenactment manipulations.
Since the KoDF dataset does not have predefined train/val/test splits, we randomly divided it into these sets, ensuring no overlap in identities. 
The number of identities for the train, validation, and test sets is 323, 40, and 40, respectively. 
We include the list of identities for the train, validation, and test splits of KoDF in our public code repository.

For cross-dataset evaluation, we adopt the Celeb-DF v2 dataset (\textbf{CDF}), DeepFakeDetection (\textbf{DFD})~\footnote{Contributing Data to Deepfake Detection Research: \url{https://blog.research.google/2019/09/contributing-data-to-deepfake-detection.html}}, and DeepFake Detection Challenge Preview (\textbf{DFDCp})~\cite{DBLP:journals/corr/abs-1910-08854}.
The CDF dataset includes 59 pairs of subjects with 590 real videos downloaded from YouTube and 5,639 high-quality fake videos generated by face swapping.
The DFD dataset is released by Google and contains 363 real videos and 3,068 fake videos featuring consenting actors. 
The DFDCp dataset contains 1,133 real videos and 4,080 fake videos produced through face swapping. 
Each dataset contains more than one video for a single identity, so an auxiliary image can be sampled from a different video corresponding to the same identity.

\subsubsection{Baseline methods}
Since our method requires sampling auxiliary images, a portion of each benchmark dataset is used for training and evaluation. 
We compare our proposed method to prior works, using identical videos for training and validation.
\begin{enumerate}
\item \textbf{LipForensics}~\cite{haliassos2021lips}: A video-level detection approach that focuses on irregularities in mouth movements.
It withstands various forms of image degradation.

\item \textbf{SBI}~\cite{DBLP:conf/cvpr/ShioharaY22}: A frame-level detection method that trains EfficientNet-B4~\cite{efficientnet} using self-blended images.
\item \textbf{TI$^2$Net}~\cite{DBLP:conf/wacv/LiuLDZY23}: A video-level detection approach that focuses on temporal identity inconsistencies. 
This method serves as the baseline for our proposed model. 
\end{enumerate}

For SBI and TI$^2$Net, we train the models using the same training data (KoDF) as our proposed method. 
For LipForensics, we utilize the model published by the authors.
The same test data is applied across all methods.

We compare our method with published values for methods whose training code is not publicly available:
\begin{enumerate}
\item \textbf{Identity Consistency Transformer} (\textbf{ICT})~\cite{dong2022protecting}: A frame-level approach focusing on identifying inconsistencies within the face’s internal and external regions for face swapping detection.
The reference-assisted ICT is indicated as ICT (ref).
\item \textbf{LRNet}~\cite{sun2021improving}: A video-level approach that analyzes the sequence of geometric features in facial videos to capture the temporal artifacts.
\item \textbf{ID-Reveal}~\cite{cozzolino2021id}: A video-level approach based on reference videos of a person and estimating a temporal embedding used as a distance metric to detect fake videos.
\item \textbf{Implicit Identity Driven} (\textbf{IID})~\cite{huang2023implicit}: A frame-level approach focused on the inconsistency between explicit and implicit facial identities in face swapping using a face recognition model.
\item \textbf{AltFreezing}~\cite{wang2023altfreezing}: A video-level approach that trains a 3D ResNet50 by alternately freezing spatial and temporal weights during training, ensuring balanced learning of both spatial and temporal artifacts.
\item \textbf{LSDA}~\cite{yan2024transcending}: A frame-level method that trains an EfficientNet-B4 by augmenting the forgery feature in latent space.
\end{enumerate}

\subsubsection{Implementation details}
We randomly sample 403 real videos and 850 fake videos from the KoDF dataset. 
For each video, we also choose a corresponding auxiliary image from a different video of the same identity.
During training, we then randomly sample 20 frame sequences consisting of 64 frames (i.e., $\ell=64$) from each video at each epoch.
Note that every video frame and auxiliary image is adjusted to align with the facial landmarks via Multitask Cascaded Convolutional Networks (MTCNN)~\cite{DBLP:journals/spl/ZhangZLQ16} and cropped to a size of $112\times112$.

For the recurrent neural networks (RNNs), we adopt the bidirectional Gated Recurrent Unit (GRU) \cite{cho2014learning}, with the dimension of the hidden state being 1,024. 
We use two fully connected layers as the classification head following the RNNs.
To prevent the model from overfitting on training data, we employ dropout layers before the RNN and the classification head, with dropout rates of 0.2 and 0.5, respectively.
We train the model over 100 epochs using the Adam optimizer~\cite{adam}, with the initial learning rate of 0.0005. 
All experiments are conducted using PyTorch~\cite{DBLP:conf/nips/PaszkeGMLBCKLGA19} on an NVIDIA Tesla V100 GPU.
Our code is publicly available here: \url{https://github.com/TaikiMiyagawa/DeepfakeDetection4eKYC}.

We use three pre-trained models as face feature extractors.
The first one is ResNet100~\cite{resnet}, pre-trained with AdaFace loss~\cite{DBLP:conf/cvpr/Kim0L22} on WebFace12M~\cite{DBLP:conf/cvpr/ZhuHDY0CZYLD021}.
It is known as one of the most advanced models for effectively processing low-quality facial images, and we primarily use this model in our experiments.
The second model is ResNet100~\cite{resnet}, pre-trained with AdaFace loss~\cite{DBLP:conf/cvpr/Kim0L22} on MS1MV2~\cite{arcface}.
The third model is ResNet100~\cite{resnet}, trained on MS1MV2~\cite{arcface} with ArcFace~\cite{arcface} loss.
The second and third models are specifically used to evaluate the impact of feature extractor selection on the accuracy and robustness of our proposed method.

\subsubsection{Evaluation metric}
We use the video-level area under the receiver operating characteristic curve (AUC).

\subsection{Robustness Assessment}

\begin{table*}[t]%
\caption{Performance declines under Gaussian blur. Declines in video-level AUCs (\%) from pristine to various levels of distortion are reported. All models are evaluated on the CDF dataset.}
\label{tab:robust_eval}
\begin{center}
\begin{tabular}{lcccccccc}
\hline
\multirow{2}{*}{Method} & \multirow{2}{*}{Training data}  & \multirow{2}{*}{Input type} & \multirow{2}{*}{AUC for pristine data}  & \multicolumn{5}{c}{AUC decline with increasing distortion level} \\
\cline{5-9}
& &  & &  1 & 2 & 3 & 4 & 5  \\
\hline
LipForensics~\cite{haliassos2021lips} & FF++ & Videos & 82.1 & 9.5 & 16.4 & 26.0 & 31.0 & 34.0 \\
SBI~\cite{DBLP:conf/cvpr/ShioharaY22} & KoDF & Frames & 83.7 & 7.6 & 11.0 & 17.5 & 24.4 & 29.0 \\
TI$^{2}$Net~\cite{DBLP:conf/wacv/LiuLDZY23} & KoDF & Videos & 54.3 & \underline{4.2} & \underline{5.7} & \underline{6.7} & \underline{5.3} & \underline{5.4}  \\
Ours & KoDF & Videos & 75.8 & \textbf{2.4} & \textbf{1.3} & \textbf{0.3} & \textbf{2.2} & \textbf{4.3} \\
\hline
\end{tabular}
\end{center}
\end{table*}

Image degradation is a common issue when images and videos are input into eKYC systems. 
We assess the robustness of existing and proposed methods by comparing the decline in AUC across varying severity levels of different image degradations.
The following operations are applied at five severity levels, as described in~\cite{deeperforensics10}: changes in saturation, changes in contrast, adding block-wise distortions, adding white Gaussian noise, Gaussian blur, and JPEG compression.

Table~\ref{tab:robust_eval} shows the AUC decline for cross-dataset evaluation using Gaussian blur as an example of image degradation. 
While our method is less accurate than SBI and LipForensics without image degradation, it experiences a much smaller AUC decline when degradation is introduced, demonstrating its robustness. 
Specifically, LipForensics and SBI show AUC declines of 34\% and 29\%, respectively, while TI$^2$Net and our method show declines of 5.4\% and 4.3\%, respectively. 
TI$^2$Net ranks second in terms of AUC decline, but its overall AUC remains consistently low. 
Robustness assessments for other image degradations are discussed below.

\subsubsection{Sensitivity to corruption severity on in-dataset evaluation}
\label{sec:robust_assess_indata}

\begin{figure*}[t]
    \centering
    \begin{minipage}[b]{0.48\linewidth}
        \centering
        \includegraphics[width=\linewidth]{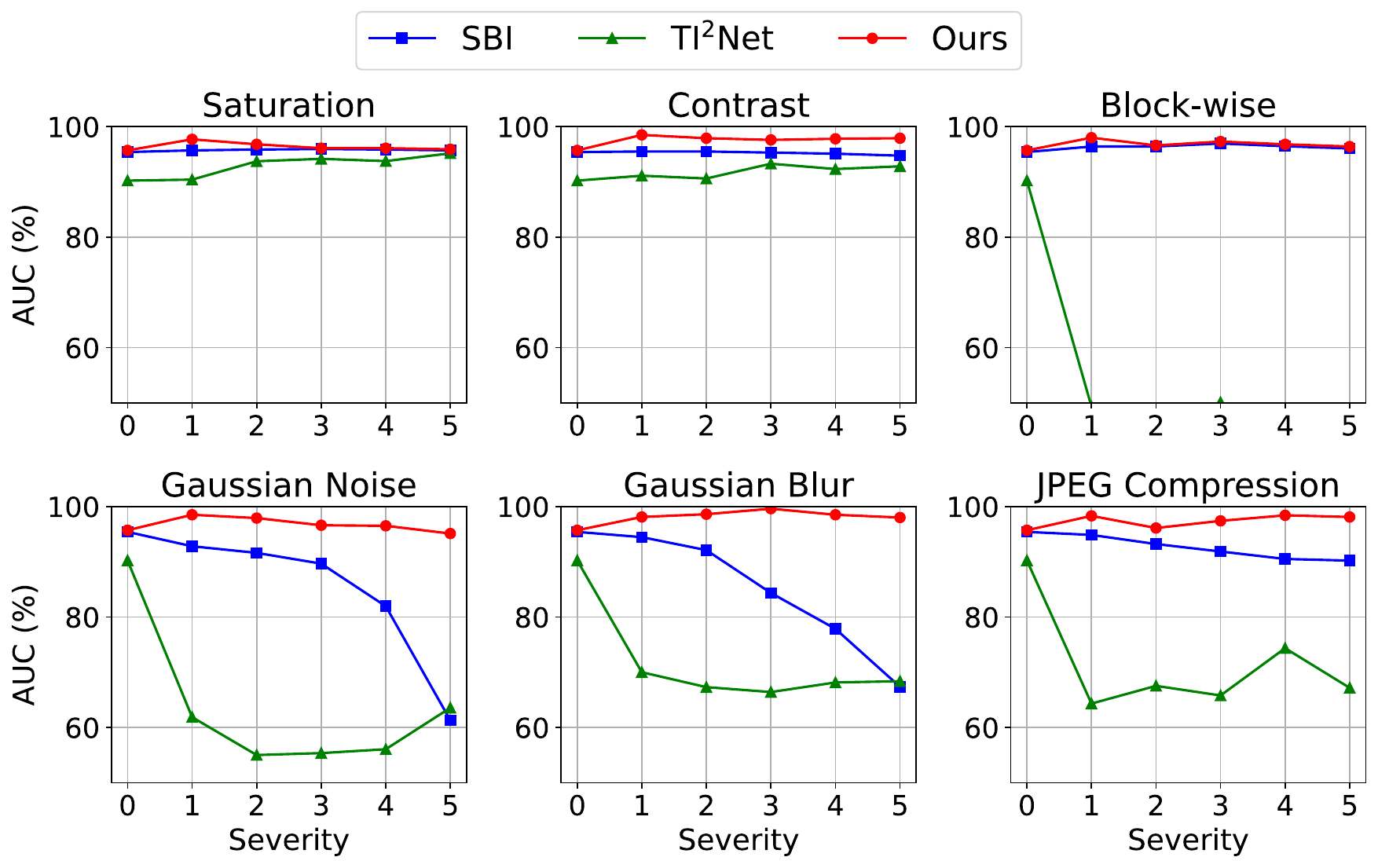}
        \caption{In-dataset evaluation of robustness against six types of image degradations. Video-level AUCs (vertical axis) are plotted as a function of severity levels (horizontal axis) for various image degradations. All models are trained and tested on the KoDF dataset. 
        }
        \label{fig:robust_kodf_both}
    \end{minipage}
    \hspace{0.4cm} 
    \begin{minipage}[b]{0.48\linewidth}
        \centering
        \includegraphics[width=\linewidth]{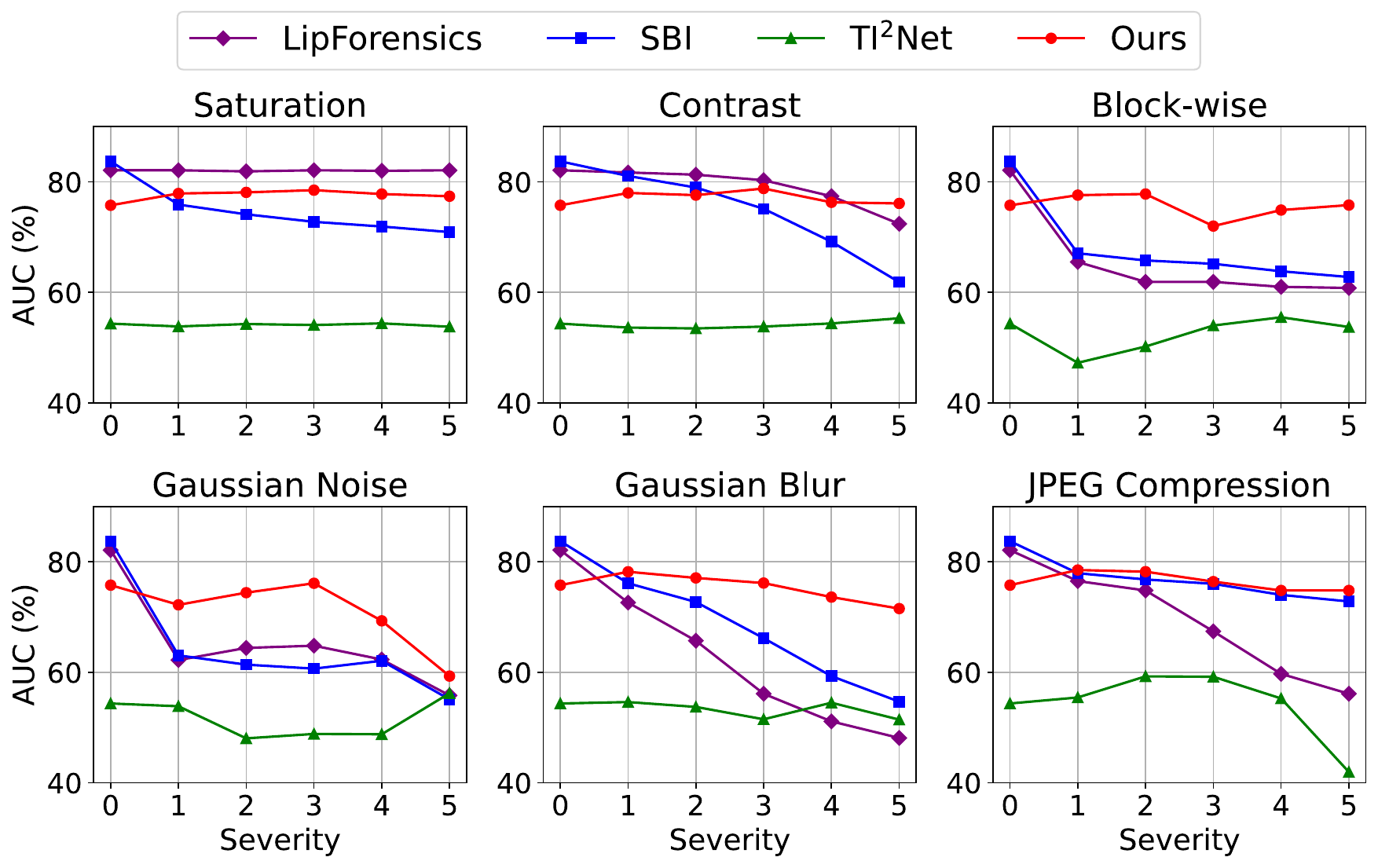}
        \caption{Cross-dataset evaluation of robustness against six types of image degradations. Video-level AUCs (vertical axis) are plotted as a function of severity levels (horizontal axis) for various image degradations.}
        \label{fig:robust_cdf_both}
    \end{minipage}
\end{figure*}

We compare the robustness of our proposed method with existing methods against six types of image degradation.
Fig.~\ref{fig:robust_kodf_both} shows the decline in AUCs due to image degradation for both existing methods and ours.
Each method is trained and evaluated on the KoDF dataset, and image transformations are applied to both real and fake data.
While AUC attenuation for saturation and contrast is minimal across all methods, SBI shows significant AUC attenuation for Gaussian noise and Gaussian blur, and TI$^2$Net exhibits substantial AUC declines for block-wise distortion, Gaussian noise, Gaussian blur, and JPEG compression.
Compared to SBI and TI$^2$Net, our method shows a smaller decline in AUC across all types of degradation, indicating the robustness of our method.

\subsubsection{Sensitivity to corruption severity on cross-dataset evaluation}
\label{sec:robust_assess_crossdata}

In Fig.~\ref{fig:robust_cdf_both}, we show the impact of the increasing severity of each corruption type on cross-dataset evaluation.
Specifically, we assess the effect of image degradation on the CDF dataset using LipForensics trained on FF++, as well as SBI, TI$^2$Net, and our method trained on KoDF. 
LipForensics and SBI exhibit similar robustness to image degradation, with minor AUC declines for saturation and contrast changes but more than 20\% degradation for other transformations. 
Although TI$^2$Net shows little AUC change under image degradation, its AUC remains around 50\%, revealing fundamental performance issues in cross-dataset evaluations. 
In most cases, our method outperforms the others, with higher AUC values indicating greater robustness to degradation. 
Since our method, which focuses on extracting identity vectors, shows less AUC decline under image degradation, identity-based deepfake detection methods appear promising for improving robustness to image degradation.

\subsection{Performance Comparison with Existing Baselines}
\label{compare_prior_works}
In this section, we comprehensively compare the experimental results of our proposed method with those reported in recent state-of-the-art methods.

\subsubsection{In-dataset Evaluation}
\label{sec:in_dataset_eval}

\begin{table}[t]
\caption{In-dataset evaluation of our proposed method and previous baselines. Each model is trained on all fake types in the KoDF dataset and tested on each fake type.}

\begin{center}
\begin{tabular}{lcccccc}
\hline
\multirow{2}{*}{Method} & \multicolumn{6}{c}{Video-level AUC (\%) for test set}\\
\cline{2-7}
 & AD & FOMM & FS & DFL & FSGAN & all  \\
\hline
TI$^2$Net~\cite{DBLP:conf/wacv/LiuLDZY23} & 91.10 & \underline{99.07} & 93.15 & \underline{92.40} & \underline{87.85} & 94.20 \\
SBI~\cite{DBLP:conf/cvpr/ShioharaY22} & \textbf{100.0} & \textbf{99.85} & \textbf{100.0} & \textbf{100.0} & 79.09 & \underline{95.40} \\
Ours & \underline{97.61} & 97.36 & \underline{97.93} & 92.20 & \textbf{96.69} & \textbf{95.70}\\
\hline
\end{tabular}
\end{center}
\label{tab:comp_prior_in}
\end{table}

We show experimental results for in-dataset evaluation in Table~\ref{tab:comp_prior_in}.
All methods are trained on the KoDF dataset and evaluated based on AUC for each KoDF fake type.
Bold and underlined values correspond to the best and the second-best values, respectively.
SBI shows high detection performance with an AUC of nearly 100\% for each fake type except for FSGAN. 
Although the proposed method performs less well than SBI for each fake type, it slightly outperforms both SBI and TI$^2$Net in the mixed evaluation (all) across all fakes types.

\subsubsection{Cross-dataset evaluation}

\begin{table*}[t]
\caption{Cross-dataset evaluation. Video-level AUCs (\%) are reported. Bold and underlined values indicate the best and the second-best results, respectively.}
\begin{center}
\footnotesize
\begin{tabular}{lcccccc}
\hline
Method & Publication & Training data & Input type & CDF & DFD & DFDCp \\
\hline
LipForensics~\cite{haliassos2021lips} & CVPR'21 & FF++ & Videos  & 82.4 & - & - \\
ID-Reveal~\cite{cozzolino2021id} & ICCV'21 & VoxCeleb2~\cite{chung2018voxceleb2} & Videos & 84.0 & - & 91.0 \\
LRNet~\cite{sun2021improving} & CVPR'21 & FF++ & Videos  & 56.9 & - & - \\
ICT~\cite{dong2022protecting} & CVPR'22 & MS1MV2 & Frames  & 85.71 & 84.13 & - \\
ICT (ref)~\cite{dong2022protecting} & CVPR'22 & MS1MV2 & Frames  & 94.43 & 93.17 & - \\
SBI~\cite{DBLP:conf/cvpr/ShioharaY22} & CVPR'22 & FF++ & Frames  & 93.18 & 97.56 & 86.15 \\
IID~\cite{huang2023implicit} & CVPR'23 & FF++ (c23) &  Frames & 83.80 & 93.92 & - \\
AltFreezing~\cite{wang2023altfreezing} & CVPR'23 & FF++ & Videos & 89.5 & 98.5 & -\\
LSDA~\cite{yan2024transcending} & CVPR'24 & FF++ (c23) & Frames & 91.1 & - & -\\
\hline
LipForensics~\cite{haliassos2021lips} & CVPR'21 & FF++ & Videos  & 82.10 & 91.59 & 74.29 \\
ICT~\cite{dong2022protecting} & CVPR'22 & MS1MV2 & Frames  & 80.58 & 86.78 & 62.94\\
ICT (ref)~\cite{dong2022protecting} & CVPR'22 & MS1MV2 & Frames  & \textbf{96.08} & \textbf{97.21} & 62.01 \\
SBI~\cite{DBLP:conf/cvpr/ShioharaY22} & CVPR'22 & KoDF & Frames & \underline{83.72} & \underline{92.71} & \underline{80.19} \\
TI$^{2}$Net~\cite{DBLP:conf/wacv/LiuLDZY23} & WACV'23 & KoDF & Videos & 64.15 & 73.59 & 61.03 \\
Ours & - & KoDF & Videos & 75.87 & 85.25 & \textbf{80.24} \\
\hline
\end{tabular}
\end{center}

\label{tab:comp_prior_cross}
\end{table*}

We compare the performance of our proposed method with existing methods in the cross-dataset setting.
Table~\ref{tab:comp_prior_cross} shows the results of the cross-dataset evaluation.
Values above the horizontal line are referenced directly from the paper, while those below are experimental values evaluated using either published models or models we trained.
SBI and ICT, which are both frame-level detection methods, perform better in cross-dataset evaluation settings such as CDF and DFD.
However, ICT struggles with the DFDCp dataset, and our proposed method outperforms all the other methods. 
The DFDCp dataset contains videos that have been post-processed by applying image degradation that reduces frames per second (FPS), resolution, and encoding quality. 
These results on the DFDCp dataset show the robustness of our method to input image degradation.

\subsection{Experiments on Various Face Feature Extractors}
\label{sec:var_feature_extractors}
\subsubsection{Sensitivity to corruption severity on in-dataset evaluation}

\begin{table*}[t]
\caption{Performance comparison across various face feature extractors. Video-level AUCs (\%) and face verification accuracy are reported.}
\label{tab:abs_id_enc_in}

\begin{center}
\begin{tabular}{l|cccccc|ccc|c}
\hline
\multirow{2}{*}{Architecture} & \multicolumn{6}{|c}{In-dataset AUC (\%)} & \multicolumn{3}{|c|}{Cross-dataset AUC (\%)} & \multirow{2}{*}{Verification accuracy}\\
\cline{2-7}\cline{7-10}
 & AD & FOMM & FS & DFL & FSGAN & All & CDF & DFD & DFDCp \\
\hline
MS1MV2/ResNet100/ArcFace & 78.21 & 83.51 & 76.61 & 86.24 & 88.98 & 81.43 & 50.75 & 56.21 & 58.18 & 57.35\\
MS1MV2/ResNet100/AdaFace & \underline{93.28} & \underline{96.69} & \textbf{98.43} & \underline{92.11} & \underline{92.52} & \underline{93.74} & \underline{69.14} & \underline{83.83} & \textbf{82.99} & \underline{65.26}\\
WebFace12M/ResNet100/AdaFace & \textbf{96.51} & \textbf{96.83} & \underline{97.50} & \textbf{92.91} & \textbf{96.76} & \textbf{95.35} & \textbf{75.87} & \textbf{85.25} & \underline{80.24} & \textbf{71.35}\\
\hline
\end{tabular}
\end{center}
\end{table*}

\begin{figure*}[t]
    \centering
    \begin{minipage}[b]{0.48\linewidth}
        \centering
        \includegraphics[width=\linewidth]{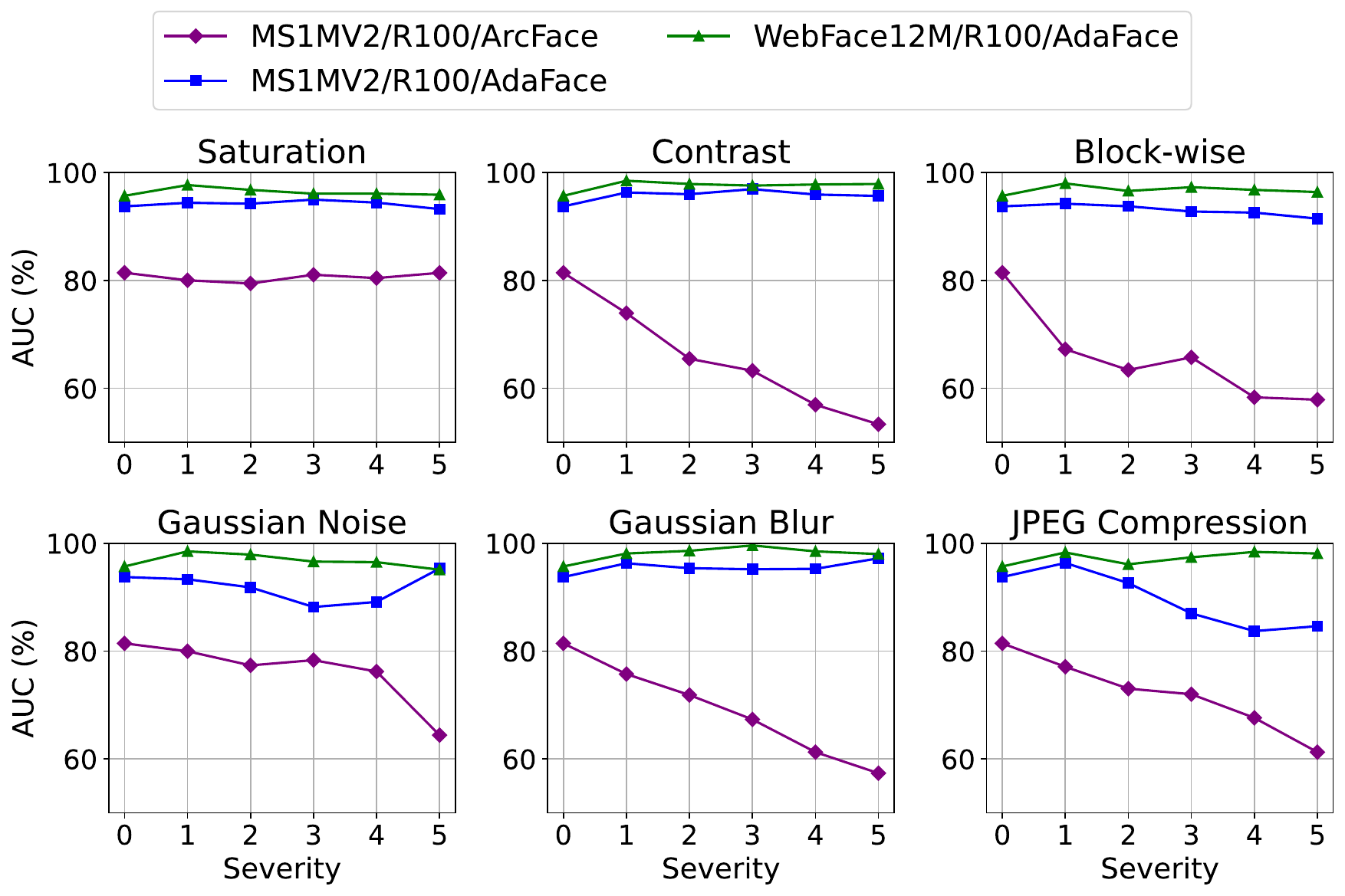}
        \caption{Robustness evaluation against six types of image degradation. Video-level AUCs (\%) (vertical axis) are plotted as a function of severity levels (horizontal axis). All models are trained and evaluated on the KoDF dataset.}
\label{fig:abs_id_enc_indata}
    \end{minipage}
    \hspace{0.4cm} 
    \begin{minipage}[b]{0.48\linewidth}
        \centering
        
        \includegraphics[width=\linewidth]{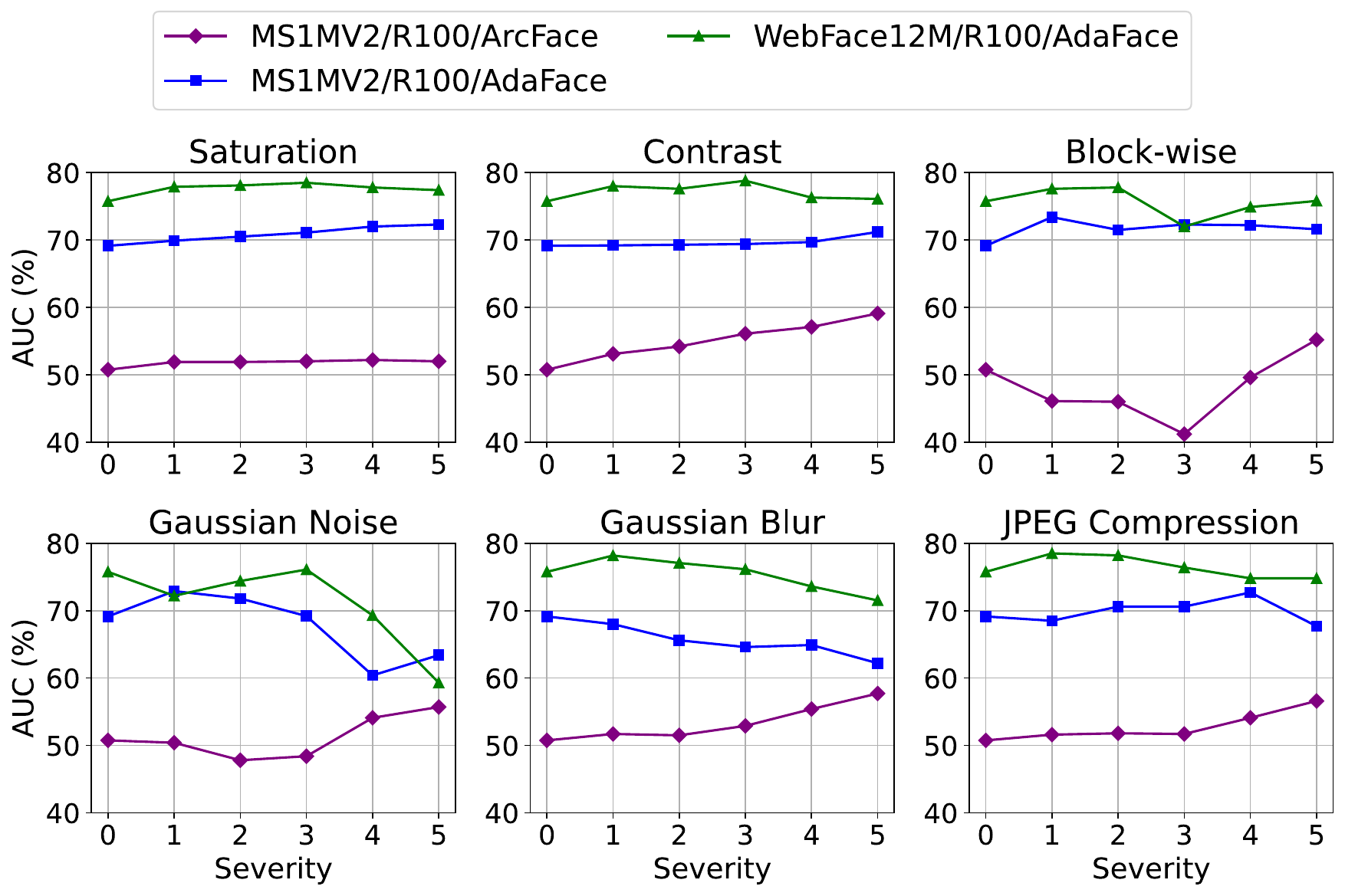}
        \caption{Robustness evaluation against six forms of image degradation. Video-level AUCs (\%) (vertical axis) are plotted as a function of severity levels (horizontal axis). All models are trained on the KoDF dataset and evaluated on the CDF dataset.}
        \label{fig:abs_id_enc_cross}
    \end{minipage}
\end{figure*}

Here, we evaluate the impact of face feature extractors. 
We compare three face feature extractors: a large-scale model using ResNet100 architecture trained with AdaFace loss on the Webface12M~\cite{DBLP:conf/cvpr/ZhuHDY0CZYLD021} dataset, a mid-scale model using ResNet100 trained with AdaFace loss on the MS1MV2 dataset, and a small-scale model using ResNet100 trained with ArcFace loss on the MS1MV2 dataset.
The accuracy comparison is shown in Table~\ref{tab:abs_id_enc_in}. 
The robustness against image distortions on the in-dataset setting is shown in Fig.~\ref{fig:abs_id_enc_indata}. 
The performance of each face feature extractor in terms of the identification accuracy of face recognition on the low-quality images is also shown, which is measured by the closed-set rank retrieval (Rank-1) in the IJB-S dataset~\cite{kalka2018ijb}.

Experimental results indicate a positive correlation between the authentication accuracy of face feature extractors and deepfake detection performance. Detectors equipped with more accurate feature extractors achieve higher accuracy. 
Table~\ref{tab:abs_id_enc_in} shows a 13.92\% increase in AUC across all fakes in the KoDF in-dataset evaluation when comparing large-scale (71.35\%) and small-scale (57.35\%) models.

Additionally, higher authentication accuracy on low-quality images enhances robustness in deepfake detection. 
Fig.~\ref{fig:abs_id_enc_indata} illustrates significant variability in AUC for the detector using the small-scale model as image quality degrades, whereas detectors with large-scale and mid-scale models maintain stable AUCs.

Furthermore, comparisons between small-scale and mid-scale models highlight that differences in training loss functions significantly influence both deepfake detection performance and robustness to image degradation.

\subsubsection{Sensitivity to corruption severity on cross-dataset evaluation}

We evaluate the effect of face feature extractors in the cross-dataset setting. 
The results of evaluating robustness against image distortion are shown in Fig.~\ref{fig:abs_id_enc_cross}. 
In a similar trend to the in-dataset evaluation, we can see that more robust face recognition leads to more robust deepfake detection, except in the case of the highest severity level of Gaussian noise.

\subsection{Experiments on Various Embedding Types}

\begin{table}[t]
\caption{Performance evaluation across various embedding types. Video-level AUCs are reported.}
\label{tab:abs_aux_img}

\begin{center}
\begin{tabular}{lcccccc}
\hline
\multirow{2}{*}{Embedding} & \multicolumn{6}{c}{Test set AUC(\%)}\\
\cline{2-7}
 & AD & FOMM & FS & DFL & FSGAN & All  \\
\hline
$D_{\rm tmp}$ & 94.44 & \textbf{99.14} & \underline{93.55} & 91.09 & 91.19 & 94.22 \\
$D_{\rm aux}$ & \textbf{98.18} & 96.59 & 90.37 & \underline{92.55} & \underline{96.00} & \underline{95.28} \\
$D_{\rm cat}$ & \underline{96.51} & \underline{96.83} & \textbf{97.50} & \textbf{92.91} & \textbf{96.67} & \textbf{95.35}\\
\hline
\end{tabular}
\end{center}
\end{table}

To evaluate the effect of the form of the features input to the RNN on the performance, we compare the performance when the model is trained with only TDC, only ADC, and their combination. 
Note that the model trained with only TDC corresponds to TI$^2$Net~\cite{DBLP:conf/wacv/LiuLDZY23} and all models' patterns train the identity vectors extracted by ResNet100 architecture trained with AdaFace loss on WebFace12M.

The results are shown in Table~\ref{tab:abs_aux_img}.
The model trained solely with TDC performs relatively well for face swapping, while the one trained solely with ADC excels in face reenactment. 
Our model, using both TDC and ADC, inherits their strengths and performs well in face swapping and face reenactment.

\subsection{Experiments on the Impact in Sampling Types}

\begin{table}[t]
\caption{Performance evaluation across various sampling types. Video-level AUCs are reported.}
\label{tab:abs_sample_type}

\begin{center}
\begin{tabular}{lccc}
\hline
\multirow{2}{*}{Sampling type} & \multicolumn{3}{c}{Cross-dataset set AUC(\%)}\\
\cline{2-4}
 & CDF & DFD & DFDCp \\
\hline
Sliding window & \textbf{75.87} & 85.25 & 80.24 \\
Random  & 74.19 & \textbf{94.21} & \textbf{91.17} \\
\hline
\end{tabular}
\end{center}
\end{table}

We compare the model's performance based on different input sequence sampling methods for the RNN. 
Sliding window sampling selects frames sequentially from the video's time series, while random sampling selects frames randomly.
Both models are trained on the KoDF dataset, and Table~\ref{tab:abs_sample_type} presents the cross-dataset evaluation results for both sampling methods. Bold values indicate the best performance.

From Table~\ref{tab:abs_sample_type}, we observe that the model using sliding window sampling shows slightly better detection performance than the model using random sampling in the CDF dataset. 
However, the model using random sampling outperforms that using sliding window sampling in the DFD and DFDCp datasets, with an 8.96\% improvement in AUC for the DFD dataset and a 10.93\% improvement in AUC for the DFDCp dataset. 
These results suggest that the sliding window sampling model may overfit the identity vector dynamics in the training data, whereas random sampling demonstrates better generalization performance.

\subsection{Experiments on the Impact of Sequence Length}

\begin{table}[t]
\caption{Performance evaluation across various sequence lengths. Video-level AUCs are reported.}
\label{tab:abs_seq_len}

\begin{center}
\begin{tabular}{cccc}
\hline
\multirow{2}{*}{Sequence length} & \multicolumn{3}{c}{Cross-dataset set AUC(\%)}\\
\cline{2-4}
 & CDF & DFC & DFDCp  \\
\hline
64 & 75.87 & \textbf{92.58} &  \underline{90.59} \\
48 & 78.36 & 91.30 &  85.27 \\
32 & 76.88 & \underline{91.46} &  88.29 \\
16 & \textbf{82.69} & 89.75 &  85.82 \\
8  & 80.08 & 89.16 &  89.22 \\
4  & \underline{80.54} & 90.17 &  \textbf{91.00} \\
2  & 79.32 & 89.11 &  83.31 \\
\hline
\end{tabular}
\end{center}
\end{table}

We compare the model's performance based on variations in the sequence length of the identity vector input to the RNN. 
We tested sequence lengths of 64, 48, 32, 16, 8, 4, and 2 with all input sequences selected using sliding window sampling. 
All models are trained on the KoDF dataset, and Table~\ref{tab:abs_seq_len} presents the cross-dataset evaluation results for each sequence length. 
Bold and underlined values correspond to the best and second-best results, respectively.

It is observed that longer sequence lengths do not necessarily lead to better detection performance. 
For the CDF dataset, models with sequence lengths of 16 and 4 show good detection performance, with a 6.87\% difference in AUC between the best and worst-performing models. 
For the DFD dataset, models with sequence lengths of 64 and 32 perform well, with a 3.47\% difference in AUC between the best and worst results. 
For the DFDCp dataset, models with sequence lengths of 4 and 64 exhibit good detection performance, with a 7.69\% difference in AUC between the best and worst results.

\section{Discussion}

The three significant components of our model (auxiliary image usage, foundation model usage, and temporal information integration) reflect a minimal yet effective design tailored to achieve robust deepfake detection in eKYC systems. While we could have added more complex structures, we adhered
to the principle of Occam's razor, valuing simplicity as a design philosophy.

\textbf{Comparison with TI$^2$Net.}
Our model introduces ADC to leverage the registered image in eKYC, a perspective overlooked by TI$^2$Net.
As shown in Table~\ref{tab:abs_aux_img}, the ADC alone surpasses competing methods on four of six datasets. When combined with TDC, the full model detects both face swapping and reenactment deepfakes with superior robustness. These results demonstrate that the registered image is a valuable cue for eKYC deepfake detection, a point previously overlooked.

\textbf{Comparison with ID-Reveal.}
ID-Reveal derives identity vectors from 3D Morphable Model-based reference videos~\cite{blanz1999amorphable}, whereas we extract them from a large-scale foundation model, aligning with the still-image enrollment typical of eKYC. 
This design also allows our system to inherit ongoing improvements in foundation models.
Our experiments confirm that larger-scale pre-training delivers higher accuracy and robustness, particularly under image degradations and cross-dataset shifts.

\section{Conclusions}

We propose a deepfake detection model optimized for eKYC that leverages the registered image and the dynamics of face identity vectors extracted using a high-performance, pre-trained face recognition model. 
Our experiments demonstrate that this approach improves the comprehensive detection of both face swapping and face reenactment, as well as its robustness to image degradation. 
Furthermore, we show that using a higher-performance face recognition model further enhances robustness against image degradation.

\section*{Ethical Impact Statement}

This study presents a deepfake detection algorithm specifically designed to enhance the security of electronic Know Your Customer (eKYC) systems against impersonation threats via face swapping and face reenactment techniques. 
As this work utilizes publicly available datasets such as KoDF and Celeb-DF v2, no new data collection involving human subjects was conducted, and thus, ethical board approval was not required. The data used aligns with open data ethics, and no personally identifiable information or sensitive individual data was generated or modified.

Our work addresses the misuse of deepfake technology by improving detection mechanisms, contributing to the prevention of identity fraud in digital authentication systems. 
Our research aims to limit negative applications of deepfake technology, particularly within sectors where identity verification is critical, such as finance and e-commerce.

Risk-mitigation strategies are embedded within our design to ensure the model's robustness across various real-world noise factors, thereby reducing risks associated with deepfake-enabled identity fraud in eKYC applications. 
We have evaluated generalizability to minimize overfitting risks that could otherwise impact the system’s utility in real-world applications.

{\small
\bibliographystyle{ieee}
\bibliography{egbib}%

\begin{thebibliography}{10}\itemsep=-1pt

\bibitem{afchar2018mesonet}
D.~Afchar, V.~Nozick, J.~Yamagishi, and I.~Echizen.
\newblock Meso{N}et: a compact facial video forgery detection network.
\newblock In {\em WIFS}, pages 1--7, 2018.

\bibitem{amerini2019deepfake}
I.~Amerini, L.~Galteri, R.~Caldelli, and A.~Del~Bimbo.
\newblock Deepfake video detection through optical flow based {CNN}.
\newblock In {\em ICCV workshops}, 2019.

\bibitem{DBLP:conf/bmvc/BalntasRPM16}
V.~Balntas, E.~Riba, D.~Ponsa, and K.~Mikolajczyk.
\newblock Learning local feature descriptors with triplets and shallow convolutional neural networks.
\newblock In {\em BMVC}, volume~1, page~3, 2016.

\bibitem{bitouk2008face}
D.~Bitouk, N.~Kumar, S.~Dhillon, P.~Belhumeur, and S.~K. Nayar.
\newblock Face swapping: automatically replacing faces in photographs.
\newblock In {\em ACM SIGGRAPH}, pages 1--8. 2008.

\bibitem{blanz1999amorphable}
V.~Blanz and T.~Vetter.
\newblock A morphable model for the synthesis of 3d faces.
\newblock In {\em ACM SIGGRAPH}, pages 187--194. 1999.

\bibitem{cho2014learning}
K.~Cho, B.~Van~Merri{\"e}nboer, C.~Gulcehre, D.~Bahdanau, F.~Bougares, H.~Schwenk, and Y.~Bengio.
\newblock Learning phrase representations using rnn encoder-decoder for statistical machine translation.
\newblock {\em arXiv:1406.1078}, 2014.

\bibitem{chung2018voxceleb2}
J.~S. Chung, A.~Nagrani, and A.~Zisserman.
\newblock Voxceleb2: Deep speaker recognition.
\newblock {\em arXiv:1806.05622}, 2018.

\bibitem{cozzolino2021id}
D.~Cozzolino, A.~R{\"o}ssler, J.~Thies, M.~Nie{\ss}ner, and L.~Verdoliva.
\newblock {ID}-reveal: Identity-aware deepfake video detection.
\newblock In {\em ICCV}, pages 15108--15117, 2021.

\bibitem{arcface}
J.~Deng, J.~Guo, N.~Xue, and S.~Zafeiriou.
\newblock Arc{F}ace: Additive angular margin loss for deep face recognition.
\newblock In {\em CVPR}, pages 4690--4699, 2019.

\bibitem{do2021potential}
T.-L. Do, M.-K. Tran, H.~H. Nguyen, and M.-T. Tran.
\newblock Potential threat of face swapping to ekyc with face registration and augmented solution with deepfake detection.
\newblock In {\em FDSE}, pages 293--307, 2021.

\bibitem{detect_deepfake_on_ekyc2}
T.-L. Do, M.-K. Tran, H.~H. Nguyen, and M.-T. Tran.
\newblock Potential threat of face swapping to ekyc with face registration and augmented solution with deepfake detection.
\newblock In {\em FDSE}, pages 293--307, 2021.

\bibitem{do2022potential}
T.-L. Do, M.-K. Tran, H.~H. Nguyen, and M.-T. Tran.
\newblock Potential attacks of deepfake on ekyc systems and remedy for ekyc with deepfake detection using two-stream network of facial appearance and motion features.
\newblock {\em SN Computer Science}, 3(6):464, 2022.

\bibitem{detect_deepfake_on_ekyc}
T.-L. Do, M.-K. Tran, H.~H. Nguyen, and M.-T. Tran.
\newblock Potential attacks of deepfake on ekyc systems and remedy for ekyc with deepfake detection using two-stream network of facial appearance and motion features.
\newblock {\em SN Computer Science}, 3(6):464, 2022.

\bibitem{DBLP:journals/corr/abs-1910-08854}
B.~Dolhansky, R.~Howes, B.~Pflaum, N.~Baram, and C.~C. Ferrer.
\newblock The deepfake detection challenge ({DFDC}) preview dataset.
\newblock {\em arXiv:1910.08854}, 2019.

\bibitem{dong2022protecting}
X.~Dong, J.~Bao, D.~Chen, T.~Zhang, W.~Zhang, N.~Yu, D.~Chen, F.~Wen, and B.~Guo.
\newblock Protecting celebrities from deepfake with identity consistency transformer.
\newblock In {\em CVPR}, pages 9468--9478, 2022.

\bibitem{felouat2024ekyc}
H.~Felouat, H.~H. Nguyen, T.-N. Le, J.~Yamagishi, and I.~Echizen.
\newblock ekyc-df: A large-scale deepfake dataset for developing and evaluating ekyc systems.
\newblock {\em IEEE Access}, 2024.

\bibitem{guera2018deepfake}
D.~G{\"u}era and E.~J. Delp.
\newblock Deepfake video detection using recurrent neural networks.
\newblock In {\em AVSS}, pages 1--6, 2018.

\bibitem{haliassos2021lips}
A.~Haliassos, K.~Vougioukas, S.~Petridis, and M.~Pantic.
\newblock Lips don't lie: A generalisable and robust approach to face forgery detection.
\newblock In {\em CVPR}, pages 5039--5049, 2021.

\bibitem{resnet}
K.~He, X.~Zhang, S.~Ren, and J.~Sun.
\newblock Deep residual learning for image recognition.
\newblock In {\em CVPR}, pages 770--778, 2016.

\bibitem{huang2023implicit}
B.~Huang, Z.~Wang, J.~Yang, J.~Ai, Q.~Zou, Q.~Wang, and D.~Ye.
\newblock Implicit identity driven deepfake face swapping detection.
\newblock In {\em CVPR}, pages 4490--4499, 2023.

\bibitem{deeperforensics10}
L.~Jiang, R.~Li, W.~Wu, C.~Qian, and C.~C. Loy.
\newblock Deeper{F}orensics-1.0: A large-scale dataset for real-world face forgery detection.
\newblock In {\em CVPR}, pages 2889--2898, 2020.

\bibitem{kalka2018ijb}
N.~D. Kalka, B.~Maze, J.~A. Duncan, K.~O’Connor, S.~Elliott, K.~Hebert, J.~Bryan, and A.~K. Jain.
\newblock Ijb-s: Iarpa janus surveillance video benchmark.
\newblock In {\em BTAS}, pages 1--9, 2018.

\bibitem{kim2018deep}
H.~Kim, P.~Garrido, A.~Tewari, W.~Xu, J.~Thies, M.~Niessner, P.~P{\'e}rez, C.~Richardt, M.~Zollh{\"o}fer, and C.~Theobalt.
\newblock Deep video portraits.
\newblock {\em ACM TOG}, 37(4):1--14, 2018.

\bibitem{DBLP:conf/cvpr/Kim0L22}
M.~Kim, A.~K. Jain, and X.~Liu.
\newblock Ada{F}ace: Quality adaptive margin for face recognition.
\newblock In {\em CVPR}, pages 18750--18759, 2022.

\bibitem{adam}
D.~P. Kingma and J.~Ba.
\newblock Adam: A method for stochastic optimization.
\newblock {\em arXiv:1412.6980}, 2014.

\bibitem{korshunova2017fast}
I.~Korshunova, W.~Shi, J.~Dambre, and L.~Theis.
\newblock Fast face-swap using convolutional neural networks.
\newblock In {\em ICCV}, pages 3677--3685, 2017.

\bibitem{DBLP:conf/iccv/KwonYNPC21}
P.~Kwon, J.~You, G.~Nam, S.~Park, and G.~Chae.
\newblock Kodf: {A} large-scale korean deepfake detection dataset.
\newblock In {\em ICCV}, pages 10744--10753, 2021.

\bibitem{li2019faceshifter}
L.~Li, J.~Bao, H.~Yang, D.~Chen, and F.~Wen.
\newblock Face{S}hifter: Towards high fidelity and occlusion aware face swapping.
\newblock {\em arXiv:1912.13457}, 2019.

\bibitem{li2020face}
L.~Li, J.~Bao, T.~Zhang, H.~Yang, D.~Chen, F.~Wen, and B.~Guo.
\newblock Face {X}-ray for more general face forgery detection.
\newblock In {\em CVPR}, pages 5001--5010, 2020.

\bibitem{li20233d}
Y.~Li, C.~Ma, Y.~Yan, W.~Zhu, and X.~Yang.
\newblock 3{D}-aware face swapping.
\newblock In {\em CVPR}, pages 12705--12714, 2023.

\bibitem{DBLP:conf/cvpr/LiYSQL20}
Y.~Li, X.~Yang, P.~Sun, H.~Qi, and S.~Lyu.
\newblock Celeb-{DF}: {A} large-scale challenging dataset for deepfake forensics.
\newblock In {\em CVPR}, pages 3207--3216, 2020.

\bibitem{DBLP:conf/wacv/LiuLDZY23}
B.~Liu, B.~Liu, M.~Ding, T.~Zhu, and X.~Yu.
\newblock {TI2N}et: temporal identity inconsistency network for deepfake detection.
\newblock In {\em WACV}, pages 4691--4700, 2023.

\bibitem{DBLP:journals/pr/LiuPGCZZ23}
K.~Liu, I.~Perov, D.~Gao, N.~Chervoniy, W.~Zhou, and W.~Zhang.
\newblock Deep{F}ace{L}ab: Integrated, flexible and extensible face-swapping framework.
\newblock {\em Pattern Recognition}, 141:109628, 2023.

\bibitem{nguyen2019use}
H.~H. Nguyen, J.~Yamagishi, and I.~Echizen.
\newblock Use of a capsule network to detect fake images and videos.
\newblock {\em arXiv:1910.12467}, 2019.

\bibitem{DBLP:conf/iccv/NirkinKH19}
Y.~Nirkin, Y.~Keller, and T.~Hassner.
\newblock {FSGAN}: Subject agnostic face swapping and reenactment.
\newblock In {\em ICCV}, pages 7184--7193, 2019.

\bibitem{DBLP:conf/nips/PaszkeGMLBCKLGA19}
A.~Paszke, S.~Gross, F.~Massa, A.~Lerer, J.~Bradbury, G.~Chanan, T.~Killeen, Z.~Lin, N.~Gimelshein, L.~Antiga, et~al.
\newblock Pytorch: An imperative style, high-performance deep learning library.
\newblock volume~32, 2019.

\bibitem{DBLP:conf/mm/PrajwalMNJ20}
K.~Prajwal, R.~Mukhopadhyay, V.~P. Namboodiri, and C.~Jawahar.
\newblock A lip sync expert is all you need for speech to lip generation in the wild.
\newblock In {\em {ACM MM}}, pages 484--492, 2020.

\bibitem{ramachandran2021experimental}
S.~Ramachandran, A.~V. Nadimpalli, and A.~Rattani.
\newblock An experimental evaluation on deepfake detection using deep face recognition.
\newblock In {\em ICCST}, pages 1--6, 2021.

\bibitem{ff++}
A.~Rossler, D.~Cozzolino, L.~Verdoliva, C.~Riess, J.~Thies, and M.~Nie{\ss}ner.
\newblock Face{F}orensics++: Learning to detect manipulated facial images.
\newblock In {\em ICCV}, pages 1--11, 2019.

\bibitem{sabir2019recurrent}
E.~Sabir, J.~Cheng, A.~Jaiswal, W.~AbdAlmageed, I.~Masi, and P.~Natarajan.
\newblock Recurrent convolutional strategies for face manipulation detection in videos.
\newblock {\em Interfaces (GUI)}, 3(1):80--87, 2019.

\bibitem{DBLP:conf/cvpr/ShioharaY22}
K.~Shiohara and T.~Yamasaki.
\newblock Detecting deepfakes with self-blended images.
\newblock In {\em CVPR}, pages 18720--18729, 2022.

\bibitem{shiohara2023blendface}
K.~Shiohara, X.~Yang, and T.~Taketomi.
\newblock Blendface: Re-designing identity encoders for face-swapping.
\newblock In {\em ICCV}, pages 7634--7644, 2023.

\bibitem{DBLP:conf/nips/SiarohinLT0S19}
A.~Siarohin, S.~Lathuilière, S.~Tulyakov, E.~Ricci, and N.~Sebe.
\newblock {F}irst {O}rder {M}otion {M}odel for {I}mage {A}nimation.
\newblock In {\em NeurIPS}, 2019.

\bibitem{sun2021improving}
Z.~Sun, Y.~Han, Z.~Hua, N.~Ruan, and W.~Jia.
\newblock Improving the efficiency and robustness of deepfakes detection through precise geometric features.
\newblock In {\em CVPR}, pages 3609--3618, 2021.

\bibitem{efficientnet}
M.~Tan and Q.~Le.
\newblock Efficientnet: Rethinking model scaling for convolutional neural networks.
\newblock In {\em ICML}, pages 6105--6114, 2019.

\bibitem{thies2019deferred}
J.~Thies, M.~Zollh{\"o}fer, and M.~Nie{\ss}ner.
\newblock Deferred neural rendering: Image synthesis using neural textures.
\newblock {\em ACM TOG}, 38(4):1--12, 2019.

\bibitem{thies2016face2face}
J.~Thies, M.~Zollhofer, M.~Stamminger, C.~Theobalt, and M.~Nie{\ss}ner.
\newblock Face2{F}ace: Real-time face capture and reenactment of rgb videos.
\newblock In {\em CVPR}, pages 2387--2395, 2016.

\bibitem{wang2023altfreezing}
Z.~Wang, J.~Bao, W.~Zhou, W.~Wang, and H.~Li.
\newblock Altfreezing for more general video face forgery detection.
\newblock In {\em CVPR}, pages 4129--4138, 2023.

\bibitem{yan2024transcending}
Z.~Yan, Y.~Luo, S.~Lyu, Q.~Liu, and B.~Wu.
\newblock Transcending forgery specificity with latent space augmentation for generalizable deepfake detection.
\newblock In {\em CVPR}, pages 8984--8994, 2024.

\bibitem{DBLP:journals/corr/abs-2002-10137}
R.~Yi, Z.~Ye, J.~Zhang, H.~Bao, and Y.-J. Liu.
\newblock Audio-driven talking face video generation with learning-based personalized head pose.
\newblock {\em arXiv:2002.10137}, 2020.

\bibitem{DBLP:journals/spl/ZhangZLQ16}
K.~Zhang, Z.~Zhang, Z.~Li, and Y.~Qiao.
\newblock Joint face detection and alignment using multitask cascaded convolutional networks.
\newblock {\em IEEE signal processing letters}, 23(10):1499--1503, 2016.

\bibitem{zhang2023sadtalker}
W.~Zhang, X.~Cun, X.~Wang, Y.~Zhang, X.~Shen, Y.~Guo, Y.~Shan, and F.~Wang.
\newblock Sadtalker: Learning realistic 3{D} motion coefficients for stylized audio-driven single image talking face animation.
\newblock In {\em CVPR}, pages 8652--8661, 2023.

\bibitem{zhao2022thin}
J.~Zhao and H.~Zhang.
\newblock Thin-plate spline motion model for image animation.
\newblock In {\em CVPR}, pages 3657--3666, 2022.

\bibitem{zhao2021learning}
T.~Zhao, X.~Xu, M.~Xu, H.~Ding, Y.~Xiong, and W.~Xia.
\newblock Learning self-consistency for deepfake detection.
\newblock In {\em ICCV}, pages 15023--15033, 2021.

\bibitem{DBLP:conf/cvpr/ZhuHDY0CZYLD021}
Z.~Zhu, G.~Huang, J.~Deng, Y.~Ye, J.~Huang, X.~Chen, J.~Zhu, T.~Yang, J.~Lu, D.~Du, et~al.
\newblock Web{F}ace260{M}: A benchmark unveiling the power of million-scale deep face recognition.
\newblock In {\em CVPR}, pages 10492--10502, 2021.

\end{thebibliography}
}

\end{document}